\documentclass{article}
\usepackage{spconf,amsmath,graphicx}
\usepackage{url}
\usepackage{bm}
\usepackage{courier}

\usepackage[pscoord]{eso-pic} 

\newcommand{\placetextbox}[4]{
  \setbox0=\hbox{#4}
  \AddToShipoutPictureFG*{
    \if#3r
    \put(\LenToUnit{\paperwidth-#1},\LenToUnit{\paperheight-#2}){\vtop{{\null}\makebox[0pt][r]{\begin{tabular}{r}#4\end{tabular}}}}%
    \else
    \put(\LenToUnit{#1},\LenToUnit{\paperheight-#2}){\vtop{{\null}\makebox[0pt][l]{\begin{tabular}{l}#4\end{tabular}}}}%
    \fi
  }%
}%
\placetextbox{0.2cm}{0.2cm}{l}{\textcopyright 2024 IEEE. Personal use of this material is permitted. Permission from IEEE must be obtained for all other uses, in any current or future media, \\ including reprinting/republishing this material for advertising or promotional purposes, creating new collective works, for resale or redistribution \\ to servers or lists, or reuse of any copyrighted component of this work in other works.}

\title{Reading Is Believing: Revisiting Language Bottleneck Models \\ for Image Classification}
\name{Honori Udo${}^*$ and Takafumi Koshinaka\thanks{$*$ Honori Udo is now with NTT Comware Corporation.}}
\address{School of Data Science, Yokohama City University\\
22-2 Seto, Kanazawa-ku, Yokohama, 236-0027, Japan}
\begin{document}
%
\maketitle
\begin{abstract}
We revisit language bottleneck models as an approach to ensuring the explainability of deep learning models for image classification. Because of inevitable information loss incurred in the step of converting images into language, the accuracy of language bottleneck models is considered to be inferior to that of standard black-box models. Recent image captioners based on large-scale foundation models of Vision and Language, however, have the ability to accurately 
describe images in verbal detail to a degree that was previously believed to not be realistically possible.
In a task of disaster image classification, we experimentally show that a language bottleneck model that combines a modern image captioner with a pre-trained language model can achieve image classification accuracy that exceeds that of black-box models. We also demonstrate that a language bottleneck model and a black-box model may be thought to extract different features from images and that fusing the two can create a synergistic effect, resulting in even higher classification accuracy.
\end{abstract}
%

\section{Introduction}
\label{sec:intro}
Advances in deep learning have given machines the ability to accurately achieve sophisticated inferences. On the other hand, a problem with these models remains in that they are black boxes, meaning that the reasons and processes behind their inferences cannot be observed. This hinders the practical application of such tasks as medical diagnoses and autonomous driving, where a single mistake can lead to seriously harmful consequences~\cite{mitchell2019}. Techniques to visualize the inference results of such complex models as deep neural networks and to help humans understand them are collectively referred to as Explainable AI (XAI), and many methods have been proposed~\cite{LIME, SHAP, Grad-CAM}. 

One approach to XAI is to define and use a set of concepts as the basis of inference. Zhou {\it et al.}~\cite{Zhou2018} 
decomposed an image of a living room into its components, such as a sofa and a table, and visualized the contribution of each component to the inference result, {\it i.e.}, living room, along with a class activation map (CAM). Murty {\it et al.}~\cite{ExpBERT} have demonstrated that text classifiers can improve the accuracy of inference w.r.t. whether A and B are married or not by adding such explanations as ``A and B went on a honeymoon'' and ``A has a daughter with B'' to the text input. 

Concept bottlenecks~\cite{Kumar2009} are a class of models that infer individual concepts held by the target object at an intermediate stage and synthesizes them in order to obtain a final outcome, making the causal relationships between concepts and outcomes transparent, and making counterfactual simulations possible through intervention. Koh {\it et al.}~\cite{Koh2020} have proposed a concept bottleneck model that linearly combines concept predictors based on modern CNNs and have achieved accuracy comparable to that with standard black-box models in X-ray image diagnoses and bird identification, while achieving high explainability with the model. However, many concepts, such as ``wing color'' and ``beak length,'' had to be conceived by humans. Yang {\it et al.}'s recent study~\cite{Yang2023} used a large language model (LLM) to automatically generate concepts. By using a foundation model of Vision and Language, such as CLIP~\cite{CLIP}, they computed the similarity between concepts and an input image in the latent space and achieved image classification accuracy comparable to black-box models in most conditions and superior to black-box models in one-shot learning conditions. 

By way of contrast, in the context of few-shot learning, in which a model is required to recognize unseen classes that are not present in the training data, research has been conducted w.r.t. using language explanations for image classification and complementing the lack of training data with linguistic knowledge. Mu {\it et al.}~\cite{Mu2020} used an image dataset with linguistic explanations (captions) in addition to class labels and have shown that better image feature representations can be obtained in the scenario of few-shot learning by training an image classification model with image captioning as an auxiliary task. In a similar framework, Afham {\it et al.}~\cite{Afham2021} have shown the effectiveness of duplicating (bi-directionalizing) the image captioning model for few-shot image classification. Andreas {\it et al.}~\cite{Andreas2018} have proposed a language bottleneck model that uses an image captioning model as a feature extractor and classifies images on the basis of language explanations alone. Nishida {\it et al.}~\cite{Nishida2022} have pointed out the problem with language bottleneck models that information is lost when converting images into language, and they have proposed an architecture that integrates a standard image classification model and a language bottleneck model, showing improved image classification accuracy. They have also shown the possibility of collaboration between machines and humans through a language input/output interface, which is a particularly interesting element in their work. 

Language bottleneck models are generally regarded as having lower image classification accuracy than standard (black-box) image classification models because of the information loss resulting from the verbalization of images. While Mu {\it et al.}'s argument~\cite{Mu2020} that it is better to use language explanations as a regularizer rather than a feature extractor sounds reasonable, the latest image captioning models, such as BLIP~\cite{BLIP}  and its successors~\cite{BLIP-2}, are very accurate and verbally detailed in describing images that they see. It might be possible to exploit their potential by using such well-trained language models as BERT~\cite{BERT}, though possibilities from this perspective have not yet been fully explored in research reported to date.  In this study, we evaluate the image classification accuracy of language bottleneck models in a standard task setting of image classification that is not few-shot and experimentally show that we can achieve performance superior to such powerful black-box image classifiers as ResNet~\cite{ResNet} and Vision Transformer (ViT)~\cite{ViT}.

The remainder of this paper is organized as follows.  Additional related work is presented in Section~\ref{sec:related}, and Section~\ref{sec:sysconfig} describes the configuration of our image classification system using image- and text-based classifiers combined with an image captioner.  The experimental setup and results for image- and text-based single-modal systems as well as fused multi-modal systems are presented in Section~\ref{sec:experiments}, and Section~\ref{sec:conclusion} summarizes our work.

\section{Related Work}
\label{sec:related}
Image captioning (image-to-text)~\cite{Ushiku2015}, which is addressed in this paper, was inspired by sequence-to-sequence learning in neural machine translation~\cite{seq2seq} and has shown remarkable progress employing a similar approach, one in which a context vector obtained by encoding an input image is decoded to generate descriptive text in an auto-regressive manner~\cite{Vinyals2015}.  It has a two-sided relationship with image generation (text-to-image), which has attracted public attention with the advent of Dall-E 2~\cite{Dall-E} and its followers~\cite{Imagen,Parti}.  

Image captioning has something in common with automatic speech recognition (ASR, a.k.a. speech-to-text), which has long received much research attention.  While ASR converts acoustic signals into text, image captioning converts light signals into text.  ASR extracts only linguistic information from input speech and discards such non-linguistic information as tone, emotion, and the gender or age of the speaker.  In this sense, ASR can be viewed as a kind of feature extraction.  Image captioning is similar in that it extracts certain information from an input image and discards other information.  This paper presents a first step towards clarifying what that “certain information” actually is.

In the field of speech emotion recognition, Srinivasan {\it et al.}~\cite{Srinivasan2022} employ linguistic information (text) obtained from ASR as features and show that a language bottleneck model using such linguistic information performs well and helps improve emotion recognition accuracy when combined with a conventional model that uses only acoustic information.  This may also apply in image captioning, even though the linguistic information an image has would be much less obvious than that which a speech utterance has.

\section{System Configuration}
\label{sec:sysconfig}
\begin{figure}[tbp]
\begin{center}
\includegraphics[scale=0.95]{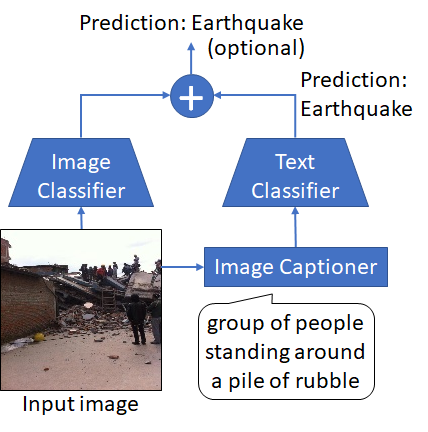}
\end{center}
\caption{System configuration: a standard image-based classifier (left-hand side) and a text-based classifier combined with an image captioner (right-hand side).}
\label{fig:system architecture}
\end{figure}

Figure~\ref{fig:system architecture} shows the configuration of the complete form of the image classification system considered in this paper.  The left-hand side shows a standard image classifier for use with a neural network, such as CNN or Transformer.  The right-hand side shows the connection of an image captioner in tandem with a text classifier for the classification of images on the basis of linguistic explanations extracted from images.

\subsection{Image and Text Classifiers}
We used pre-trained models throughout.  ResNet-50, ResNet-101~\cite{ResNet}, ViT-Base, and ViT-Large~\cite{ViT} were used for the image-based classifier, and BERT${}_{\rm BASE}$~\cite{BERT} was used for the text-based classifier~\footnote{All the models we used are available on the Hugging Face Hub: 'microsoft/resnet-50', 'microsoft/resnet-101', 'google/vit-base-patch16-224-in21k', 'google/vit-huge-patch14-224-in21k', and 'bert-base-uncased'.} after fine-tuning with the training data for the target task.  Both of those models are known to be highly competitive over a wide range of classification tasks.  Hyper-parameter settings in fine-tuning, {\it i.e.}, the learning rate, mini-batch size, and number of epochs, were set at $10^{-4}$, 128, and 30, respectively, with the Adam optimizer.  Note that we first freeze the body of the model to train the final linear layer only, and then unfreeze the body to train the entire model parameters for the last two (BERT) or three (ResNet and ViT) epochs, and this results in consistently good performance on the development set.




\subsection{Image Captioners}
We focused on the four image captioning models listed below. None of them were fine-tuned using the data for the target task (because no image description text was available for the target task), and the original models were used as it is.
\vspace{3mm}

\noindent
{\bf InceptionV3+RNN}~\cite{Attention}: This is a basic, small-scale model that encodes an input image into a vector by using InceptionV3~\cite{InceptionV3} and then decodes it to generate a caption using a recurrent neural network (GRU).  An attention mechanism is placed between the encoder and decoder, and features of individual parts of the image are selectively sent to the decoder.  The entire system was trained with the MS-COCO dataset~\cite{MS-COCO}, and we followed TensorFlow's tutorial implementation~\footnote{
\url{https://www.tensorflow.org/tutorials/text/image_captioning}.}.

\noindent
{\bf BLIP}~\cite{BLIP}: This is a foundation model that has learned a large number of images and amount of text and is applicable to a wide range of Vision and Language tasks.  When used as an image captioner, it takes the form of an encoder-decoder configuration based on the Transformer architecture.  Users can easily run the sample code (demo.ipynb) on GitHub~\footnote{\url{https://github.com/salesforce/BLIP}} to obtain captions for their own images.  BLIP is a relatively advanced, large-scale model that is capable of producing quite accurate captions.  In our experiment, ViT-Large (307M parameters) was used for the image encoder.

\noindent
{\bf BLIP-2}~\cite{BLIP-2}: This second version of BLIP can connect any pre-trained image encoder and large language model (LLM) through a flexible component referred to as Q-Former and can produce higher-quality captions from given images thanks to large-scale pre-training. In the default setting, the image encoder and LLM are, respectively, ViT-g/14~\cite{ViT-g}, which is an even larger model (1011M parameters) than ViT-Large, and Meta AI's OPT-2.7B~\cite{OPT}~\footnote{\url{https://github.com/salesforce/LAVIS/tree/main/projects/blip2}}.

\noindent
{\bf CLIP Interrogator}: Given an image, this model infers prompts for such AI image generators as Stable Diffusion and Midjourney, so as to generate similar images.
Since the text generated by CLIP Interrogator is not meant to be read by humans, it may not be considered an image captioner in the strict sense, but we tested it as a model that can generate richer text than can BLIP-2.

Although the technical specifications of CLIP Interrogator have not been published as a paper and there is no relevant literature that can be referred to, it may be presumed from the code~\footnote{\url{https://github.com/pharmapsychotic/clip-interrogator}} and its operation that it first generates a base caption using BLIP and then selects and adds phrases that match the target image from a predefined set of phrases called {\it Flavors}.  CLIP image/text encoders~\cite{CLIP} are used to measure the degree of matching between a target image and the phrases in {\it Flavors}.  {\it Flavors} contains approximately 100,000 words and phrases, including those referring to objects and entities (e.g., motorcycle, building, young woman), image styles (e.g., photo-realistic), and artist names (e.g., greg rutkowski).  We used the code released by the developer (clip\_interrogator.ipynb, version 2.2).
\vspace{3mm}


\subsection{System Fusion}
As previously indicated in Figure~\ref{fig:system architecture}, we fuse the outputs of an image-based classifier with those of a text-based classifier to improve classification accuracy.  Possible fusion methods include feature-level fusion (early fusion), which inputs the hidden layer states of each classifier into another neural-network classifier, and score-level fusion (late fusion), which averages the classification results of each classifier.  Here we chose the latter for simplicity.  Suppose, for example, that the number of classes is $C$ and the output of the image/text-based classifiers normalized by the softmax function are
$\bm{y}^{\left(I\right)}=\left(y_1^{\left(I\right)},\cdots,y_C^{\left(I\right)}\right)$ and
$\bm{y}^{\left(T\right)}=\left(y_1^{\left(T\right)},\cdots,y_C^{\left(T\right)}\right)$,
respectively. The classification results obtained with score-level fusion may then be calculated as
$\bm{y}=\left(1-w\right)\bm{y}^{\left(I\right)}+w\bm{y}^{\left(T\right)}$, where $0 \le w \le 1$ is the weight coefficient for the text-based classifier.

\section{Experiments}
\label{sec:experiments}
\begin{figure}[tbp]
\begin{center}
\includegraphics[scale=0.45]{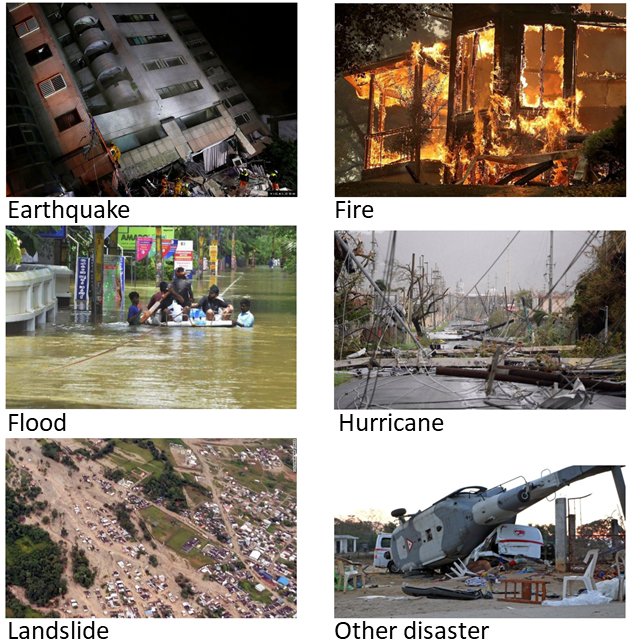}
\end{center}
\caption{Example images for different disaster types included in the CrisisNLP dataset (cited from \cite{CrisisNLP}). There is another type, referred to as "not disaster," which is not shown here.}
\label{fig:CrisisNLP}
\end{figure}

We used the CrisisNLP dataset~\cite{CrisisNLP, CrisisMMD}, which is a collection of natural disaster images shared on such social media as Twitter (currently rebranded as X)~\footnote{\url{https://crisisnlp.qcri.org/crisis-image-datasets-asonam20}}.  The dataset provides four image classification tasks, for each of which training (Train), development (Dev), and test (Test) data partitions are defined.  Among them, we focus on two tasks: 
1) ``Disaster types,'' for predicting types of disasters, such as earthquakes, floods, etc.;
2)  ``Damage severity,'' for predicting the degree of damage caused by a disaster in terms of three levels: ``none or little,'' ``mild,'' and ``severe.'' (See Table~\ref{tab:dataset} and Figure~\ref{fig:CrisisNLP})

\begin{table}[t]
\caption{Number of images and classes contained in two tasks defined in the CrisisNLP dataset.}
\label{tab:dataset}
\begin{center}
\begin{tabular}{|c||c|c|} \hline 
Task    & Disaster types & Damage severity \\ \hline\hline
Train & 12,724 & 26,898 \\
Dev   &  1,574 &  2,898 \\
Test  &  3,213 &  5,100 \\ \hline
\# Classes & 7 & 3 \\ \hline
\end{tabular}
\end{center}
\end{table}

\begin{table}[tbp]
\caption{Accuracies (\%) of image-based classifiers: ResNet-50, ResNet-101, ViT-Base, ViT-Large; and text-based classifiers: BERT combined with InceptionV3+RNN (IV3+RNN), BLIP, BLIP-2, and CLIP Interrogator (CLIP-I).}
\label{tab:single modal systems}
\begin{center}
\begin{tabular}{|l|cc|} \hline
System & Disaster types & Damage severity \\ \hline\hline
{\em Image-based} & & \\
\hspace{2mm} ResNet-50  & 78.38 & 77.60 \\
\hspace{2mm} ResNet-101 & 79.71 & 77.40 \\
\hspace{2mm} ViT-Base   & {\bf 84.22} & {\bf 78.99} \\
\hspace{2mm} ViT-Large  & 82.01 & 77.56 \\ \hline
{\em Text-based}  & & \\
\hspace{2mm} IV3+RNN & 42.38 & 55.13 \\
\hspace{2mm} BLIP    & 70.55 & 72.49 \\
\hspace{2mm} BLIP-2  & 78.11 & 76.78 \\
\hspace{2mm} CLIP-I  & {\bf 85.09} & {\bf 79.94} \\ \hline
\end{tabular}
\end{center}
\end{table}

We first show the classification accuracies of single-modal systems using only an image-based classifier or a text-based classifier (Table~\ref{tab:single modal systems}).  To reduce the randomness of model parameter initialization in fine-tuning, each of those accuracies is averaged over five trials.

\begin{figure}[t]
\begin{center}
\begin{tabular}{|p{15mm}|p{56mm}|} \hline
Input & 
\raisebox{-0.97\height}{\includegraphics{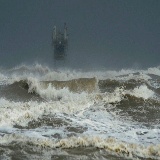}} \\ \hline
ViT-Base & $\rightarrow$ {\bf not disaster} \\ \hline
IV3+RNN & a snowboard near another wave in the water 
$\rightarrow$ {\bf flood} \\ \hline
BLIP & an oil rig in the middle of the ocean on a foggy day 
$\rightarrow$ {\bf not disaster} \\ \hline
BLIP-2 & a large wave is crashing over the ocean 
$\rightarrow$ {\bf hurricane} \\ \hline
CLIP-I & a large body of water with a boat in the distance, stormy seas, stormy sea, rough seas, tumultuous sea, rough sea, violent stormy waters, storm at sea, rough water, apocalyptic tumultuous sea, a violent storm at sea, towering waves, sea storm, in rough seas with large waves, rough seas in background, stormy wheater 
$\rightarrow$ {\bf hurricane} \\ \hline
\end{tabular}
\end{center}
\caption{Example results with an image to be classified as ``hurricane.''}
\label{fig:Example result}
\end{figure}

Regarding the image-based classifiers, the ViT models, particularly ViT-Base in ``Disaster type'' classification, outperformed ResNet, showing the strength of the Transformer architecture.  Comparing the text-based classifier with four different image captioners, we can first see that the most basic image captioner, InceptionV3+RNN (IV3+RNN), falls far short of obtaining the classification accuracy of standard image-based classifiers.  A look at the captions generated by IV3+RNN reveals that most of them are seemingly irrelevant with respect to the images, and it seems difficult to predict either the type of disaster or the degree of damage from these captions (as indicated later in Figure \ref{fig:Example result}).  By way of contrast, BLIP and BLIP-2 did generate good captions for many images.  The caption previously shown in Figure~\ref{fig:system architecture} is one actually generated by BLIP, and it notes such important elements in the image as ``people'' and ``rubble.''  The text-based classifier with BLIP consequently achieved much better accuracy.  The one with BLIP-2, based on large-scale pre-trained models, did even better and reached an accuracy level roughly comparable to that of some standard image-based classifiers.  The accuracy of the text-based classifier using CLIP Interrogator (CLIP-I) went beyond BLIP-2. Its results exceed those of ViT-Base, the best image-based classifier. This suggests that foundation models of Vision and Language trained on a large amount of image/text data would seem to be a promising option for image feature extraction.

Figure~\ref{fig:Example result} shows an example of image classification results for an image that should be classified as ``hurricane.''  As previously noted, InceptionV3+RNN (IV3+RNN), which was the most basic image captioner, produced a description that was far from the actual content of the image.  BLIP's descriptions were generally much more accurate, though in this case it focuses on the oil rig in the background and fails to catch the rough waves. What BLIP-2 generated here looks perfect. The CLIP Interrogator (CLIP-I) behaved quite differently from the others.  After beginning with a normal sentence coming from BLIP, it continued the explanation with a large number of phrases selected by CLIP.  Although those phrases may be unhelpful as specific information, we can observe some that reflect the true class, such as ``large waves'' and ``violent storm.''  On the other hand, we found CLIP Interrogator often generated completely irrelevant phrases, {\it e.g.}, ``movie poster'' and ''youtube video screenshot;'' they would, however, be good clues for Stable Diffusion to use in the generation of images.  It is assumed that the additional phrases selected by CLIP produce an effect similar to that of data augmentation.

\begin{figure}[t!]
\begin{flushleft}
(a) Disaster types
\end{flushleft}
\begin{center}
\includegraphics[width=\linewidth]{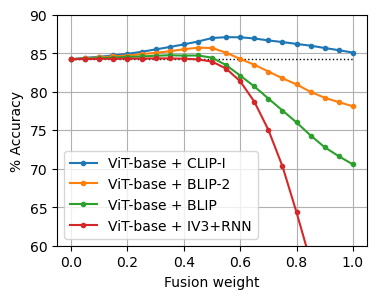}
\end{center}
\begin{flushleft}
(b) Damage severity
\end{flushleft}
\begin{center}
\includegraphics[width=\linewidth]{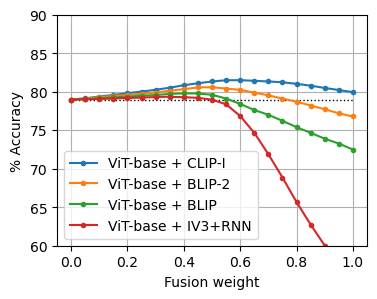}
\end{center}
\caption{Score-level fusion results using ViT-Base as the image-based classifier: Horizontal axis represents the fusion weight $w$ for the text-based classifier.  $w=0$ and $w=1$ correspond to image-based and text-based single-modal systems, respectively.}
\label{fig:Fusion}
\end{figure}

Figure~\ref{fig:Fusion} shows the results of score-level fusion that averaged the output of the image/text-based classifiers with weight $w$, where ViT-Base, the best-performing vision model, was used for the image-based classifier.  Both of the two tasks (Disaster types, Damage severity) show similar trends, {\it i.e.}, when using sufficiently good models, such as BLIP-2 and CLIP Interrogator (CLIP-I), for image captioning, classification accuracy can be improved by appropriately choosing the weight $w$.  It seems that image captioning models extract features different from vision models; in other words, they look at images from a different perspective than do vision models.  It might also be worth mentioning that a similar trend was observed when using CNN-based models, {\it i.e.} ResNet, instead of Transformer-based ones.

\section{Summary}
\label{sec:conclusion}
We have revisited language bottleneck models as an approach to ensure the explainability of deep learning models in image classification, and experimentally examined its usefulness. Because information is inevitably lost in the step of converting images into language, the accuracy of the language bottleneck model is considered to be inferior to the standard black-box model. However, image captioning based on recent large-scale foundation models of Vision and Language has the ability to describe images accurately and in verbal detail, and when combined with a well-trained language model, it achieves image classification accuracy that exceeds that of black-box models. It has also shown that the language bottleneck model and the black-box model are thought to extract different features from images, and that a synergistic effect can be obtained by integrating the two.

In our experiments, we obtained results using disaster image data, but these results strongly depend, of course, on the data. In future work, we intend to verify our results with a variety of datasets from different domains, such as CUB~\cite{CUB} (bird identification) and OAI~\cite{OAI} (X-ray imaging diagnosis). We would also hope to investigate in detail what kind of features image captioners extract from images and how such features differ from those extracted by vision models. Further, while our model integration in the experiment was in a form of simple score-level fusion, we see significant potential in exploring both feature-level fusion and knowledge distillation (teacher-student learning), as has been attempted in speech emotion recognition~\cite{Srinivasan2022}.

\section{Acknowledgment}
This work was partially supported by JSPS KAKEN Grant Numbers 21K11967 and 24K15012.

\bibliographystyle{IEEEbib}
\bibliography{main}

\begin{thebibliography}{10}

\bibitem{mitchell2019}
Melanie Mitchell,
\newblock {\em Artificial Intelligence: A Guide for Thinking Humans},
\newblock Macmillan Publishers (Farrar, Straus and Giroux), 2019.

\bibitem{LIME}
Marco~Tulio Ribeiro, Sameer Singh, and Carlos Guestrin,
\newblock ``Why should {I} trust you?: Explaining the predictions of any classifier,''
\newblock in {\em Proceedings of the 22nd ACM SIGKDD International Conference on Knowledge Discovery and Data Mining}, 2016.

\bibitem{SHAP}
Scott~M Lundberg and Su-In Lee,
\newblock ``A unified approach to interpreting model predictions,''
\newblock in {\em Advances in Neural Information Processing Systems}, 2017.

\bibitem{Grad-CAM}
Ramprasaath~R. Selvaraju, Michael Cogswell, Abhishek Das, Ramakrishna Vedantam, Devi Parikh, and Dhruv Batra,
\newblock ``Grad-{CAM}: Visual explanations from deep networks via gradient-based localization,''
\newblock in {\em IEEE International Conference on Computer Vision (ICCV)}, 2017.

\bibitem{Zhou2018}
Bolei Zhou, Yiyou Sun, David Bau, and Antonio Torralba,
\newblock ``Interpretable basis decomposition for visual explanation,''
\newblock in {\em Proceedings of the European Conference on Computer Vision (ECCV)}, 2018.

\bibitem{ExpBERT}
Shikhar Murty, Pang~Wei Koh, and Percy Liang,
\newblock ``{E}xp{BERT}: Representation engineering with natural language explanations,''
\newblock in {\em Proceedings of the 58th Annual Meeting of the Association for Computational Linguistics}, 2020.

\bibitem{Kumar2009}
Neeraj Kumar, Alexander~C. Berg, Peter~N. Belhumeur, and Shree~K. Nayar,
\newblock ``Attribute and simile classifiers for face verification,''
\newblock in {\em 2009 IEEE 12th International Conference on Computer Vision}, 2009.

\bibitem{Koh2020}
Pang~Wei Koh, Thao Nguyen, Yew~Siang Tang, Stephen Mussmann, Emma Pierson, Been Kim, and Percy Liang,
\newblock ``Concept bottleneck models,''
\newblock in {\em Proceedings of the 37th International Conference on Machine Learning}, 2020.

\bibitem{Yang2023}
Yue Yang, Artemis Panagopoulou, Shenghao Zhou, Daniel Jin, Chris Callison-Burch, and Mark Yatskar,
\newblock ``Language in a bottle: Language model guided concept bottlenecks for interpretable image classification,''
\newblock in {\em Proceedings of the IEEE/CVF Conference on Computer Vision and Pattern Recognition (CVPR)}, June 2023.

\bibitem{CLIP}
Alec Radford, Jong~Wook Kim, Chris Hallacy, Aditya Ramesh, Gabriel Goh, Sandhini Agarwal, Girish Sastry, Amanda Askell, Pamela Mishkin, Jack Clark, Gretchen Krueger, and Ilya Sutskever,
\newblock ``Learning transferable visual models from natural language supervision,'' arXiv:2103.00020, 2021.

\bibitem{Mu2020}
Jesse Mu, Percy Liang, and Noah Goodman,
\newblock ``Shaping visual representations with language for few-shot classification,''
\newblock in {\em Proceedings of the 58th Annual Meeting of the Association for Computational Linguistics}, 2020.

\bibitem{Afham2021}
Mohamed Afham~Mohamed Aflal, Salman Khan, Muhammad~Haris Khan, Muzammal Naseer, and Fahad~Shahbaz Khan,
\newblock ``Rich semantics improve few-shot learning,''
\newblock in {\em Proceedings of The 32nd British Machine Vision Conference (BMVC)}, 2021.

\bibitem{Andreas2018}
Jacob Andreas, Dan Klein, and Sergey Levine,
\newblock ``Learning with latent language,''
\newblock in {\em Proceedings of the 2018 Conference of the North {A}merican Chapter of the Association for Computational Linguistics: Human Language Technologies}, 2018.

\bibitem{Nishida2022}
Kosuke Nishida, Kyosuke Nishida, and Shuichi Nishioka,
\newblock ``Improving few-shot image classification using machine- and user-generated natural language descriptions,''
\newblock in {\em Findings of the Association for Computational Linguistics: NAACL 2022}, 2022.

\bibitem{BLIP}
Junnan Li, Dongxu Li, Caiming Xiong, and Steven Hoi,
\newblock ``{BLIP}: Bootstrapping language-image pre-training for unified vision-language understanding and generation,'' arXiv:2201.12086, 2022.

\bibitem{BLIP-2}
Junnan Li, Dongxu Li, Silvio Savarese, and Steven Hoi,
\newblock ``{BLIP}-2: Bootstrapping language-image pre-training with frozen image encoders and large language models,'' arXiv:2301.12597, 2023.

\bibitem{BERT}
Jacob Devlin, Ming-Wei Chang, Kenton Lee, and Kristina Toutanova,
\newblock ``{BERT}: Pre-training of deep bidirectional transformers for language understanding,''
\newblock in {\em Proceedings of the 2019 Conference of the North {A}merican Chapter of the Association for Computational Linguistics: Human Language Technologies}, 2019.

\bibitem{ResNet}
Kaiming He, Xiangyu Zhang, Shaoqing Ren, and Jian Sun,
\newblock ``Deep residual learning for image recognition,''
\newblock in {\em 2016 IEEE Conference on Computer Vision and Pattern Recognition (CVPR)}, 2016.

\bibitem{ViT}
Alexey Dosovitskiy, Lucas Beyer, Alexander Kolesnikov, Dirk Weissenborn, Xiaohua Zhai, Thomas Unterthiner, Mostafa Dehghani, Matthias Minderer, Georg Heigold, Sylvain Gelly, Jakob Uszkoreit, and Neil Houlsby,
\newblock ``An image is worth 16x16 words: Transformers for image recognition at scale,''
\newblock in {\em 9th International Conference on Learning Representations ({ICLR})}, 2021.

\bibitem{Ushiku2015}
Yoshitaka Ushiku, Masataka Yamaguchi, Yusuke Mukuta, and Tatsuya Harada,
\newblock ``Common subspace for model and similarity: Phrase learning for caption generation from images,''
\newblock in {\em 2015 IEEE International Conference on Computer Vision (ICCV)}, 2015.

\bibitem{seq2seq}
Ilya Sutskever, Oriol Vinyals, and Quoc~V. Le,
\newblock ``Sequence to sequence learning with neural networks,''
\newblock in {\em Proceedings of the 27th International Conference on Neural Information Processing Systems}, 2014.

\bibitem{Vinyals2015}
Oriol Vinyals, Alexander Toshev, Samy Bengio, and Dumitru Erhan,
\newblock ``Show and tell: A neural image caption generator,''
\newblock in {\em Proceedings of the IEEE Conference on Computer Vision and Pattern Recognition (CVPR)}, 2015.

\bibitem{Dall-E}
Aditya Ramesh, Prafulla Dhariwal, Alex Nichol, Casey Chu, and Mark Chen,
\newblock ``Hierarchical text-conditional image generation with {CLIP} latents,'' arXiv:2204.06125, 2022.

\bibitem{Imagen}
Chitwan Saharia, William Chan, Saurabh Saxena, Lala Li, Jay Whang, Emily~L Denton, Kamyar Ghasemipour, Raphael Gontijo~Lopes, Burcu Karagol~Ayan, Tim Salimans, Jonathan Ho, David~J Fleet, and Mohammad Norouzi,
\newblock ``Photorealistic text-to-image diffusion models with deep language understanding,''
\newblock in {\em Advances in Neural Information Processing Systems}, 2022.

\bibitem{Parti}
Jiahui Yu, Yuanzhong Xu, Jing~Yu Koh, Thang Luong, Gunjan Baid, Zirui Wang, Vijay Vasudevan, Alexander Ku, Yinfei Yang, Burcu~Karagol Ayan, Ben Hutchinson, Wei Han, Zarana Parekh, Xin Li, Han Zhang, Jason Baldridge, and Yonghui Wu,
\newblock ``Scaling autoregressive models for content-rich text-to-image generation,''
\newblock {\em Transactions on Machine Learning Research}, 2022.

\bibitem{Srinivasan2022}
Sundararajan Srinivasan, Zhaocheng Huang, and Katrin Kirchhoff,
\newblock ``Representation learning through cross-modal conditional teacher-student training for speech emotion recognition,''
\newblock in {\em 2022 IEEE International Conference on Acoustics, Speech and Signal Processing (ICASSP)}, 2022.

\bibitem{Attention}
Kelvin Xu, Jimmy Ba, Ryan Kiros, Kyunghyun Cho, Aaron Courville, Ruslan Salakhudinov, Rich Zemel, and Yoshua Bengio,
\newblock ``Show, attend and tell: Neural image caption generation with visual attention,''
\newblock in {\em Proceedings of the 32nd International Conference on Machine Learning}, Jul 2015.

\bibitem{InceptionV3}
Christian Szegedy, Vincent Vanhoucke, Sergey Ioffe, Jon Shlens, and Zbigniew Wojna,
\newblock ``Rethinking the inception architecture for computer vision,''
\newblock in {\em 2016 IEEE Conference on Computer Vision and Pattern Recognition (CVPR)}, 2016.

\bibitem{MS-COCO}
Tsung-Yi Lin, Michael Maire, Serge Belongie, James Hays, Pietro Perona, Deva Ramanan, Piotr Doll{\'a}r, and C.~Lawrence Zitnick,
\newblock ``Microsoft {COCO}: Common objects in context,''
\newblock in {\em Computer Vision -- ECCV 2014}, 2014.

\bibitem{ViT-g}
Xiaohua Zhai, Alexander Kolesnikov, Neil Houlsby, and Lucas Beyer,
\newblock ``Scaling vision transformers,''
\newblock in {\em Proceedings of the IEEE/CVF Conference on Computer Vision and Pattern Recognition (CVPR)}, 2022.

\bibitem{OPT}
Susan Zhang, Stephen Roller, Naman Goyal, Mikel Artetxe, Moya Chen, Shuohui Chen, Christopher Dewan, Mona Diab, Xian Li, Xi~Victoria Lin, Todor Mihaylov, Myle Ott, Sam Shleifer, Kurt Shuster, Daniel Simig, Punit~Singh Koura, Anjali Sridhar, Tianlu Wang, and Luke Zettlemoyer,
\newblock ``{OPT}: Open pre-trained transformer language models,'' arXiv:2205.01068, 2022.

\bibitem{CrisisNLP}
Firoj Alam, Ferda Ofli, Muhammad Imran, Tanvirul Alam, and Umair Qazi,
\newblock ``Deep learning benchmarks and datasets for social media image classification for disaster response,''
\newblock in {\em Proceedings of the 12th IEEE/ACM International Conference on Advances in Social Networks Analysis and Mining}. 2021, IEEE Press.

\bibitem{CrisisMMD}
Firoj Alam, Ferda Ofli, and Muhammad Imran,
\newblock ``Crisis{MMD}: Multimodal twitter datasets from natural disasters,''
\newblock in {\em Proceedings of the International AAAI Conference on Web and Social Media}, 2018.

\bibitem{CUB}
C.~Wah, S.~Branson, P.~Welinder, P.~Perona, and S.~Belongie,
\newblock ``Cub-200-2011,'' CaltechDATA, Apr 2022.

\bibitem{OAI}
C.~G. Peterfy, E.~Schneider, and M.~Nevitt,
\newblock ``The osteoarthritis initiative: report on the design rationale for the magnetic resonance imaging protocol for the knee,'' Osteoarthritis Cartilage, Dec 2008.

\end{thebibliography}
\end{document}